\documentclass[pdflatex,sn-mathphys-num]{sn-jnl}

\usepackage{xspace}
\usepackage{booktabs}
\usepackage{array}
\usepackage{subfigure}
\usepackage{balance}
\usepackage{graphicx}
\usepackage{epsfig}
\usepackage{amsmath,bm}
\usepackage{algorithm}
\usepackage{algorithmic}
\usepackage{epstopdf}
\usepackage{enumerate}
\usepackage{tabularx}
\usepackage{amsthm}
\usepackage{amsfonts}
\usepackage{color}

\newcommand{\ourmethod}{\texttt{ANEMONE}\xspace}
\newcommand{\ourmethodfs}{\texttt{ANEMONE-FS}\xspace}

\theoremstyle{definition}
\newtheorem{definition}{Definition}[section]

\DeclareMathAlphabet\mathbfcal{OMS}{cmsy}{b}{n}

\newcommand{\mysubsubtitle}[1]{\noindent \textbf{#1}.}

\theoremstyle{thmstyleone}%
\theoremstyle{thmstyletwo}%

\theoremstyle{thmstylethree}%

\begin{document}

\title[Article Title]{From Unsupervised to Few-shot Graph Anomaly Detection: A Multi-scale Contrastive Learning Approach}


\author[1]{\fnm{Yu} \sur{Zheng}}\email{yu.zheng@latrobe.edu.au}

\author[2]{\fnm{Ming} \sur{Jin}}\email{ming.jin@griffith.edu.au}

\author[2]{\fnm{Yixin} \sur{Liu}}\email{yixin.liu@griffith.edu.au}

\author[1]{\fnm{Lianhua} \sur{Chi}}\email{l.chi@latrobe.edu.au}
\equalcont{Corresponding Author}

\author[1]{\fnm{Khoa} \sur{Phan}}\email{k.phan@latrobe.edu.au}

\author[1]{\fnm{Yi-Ping Phoebe} \sur{Chen}}\email{phoebe.chen@latrobe.edu.au}

\affil[1]{\orgdiv{Department of Computer Science and Information Technology}, \orgname{La Trobe University}, \orgaddress{ \city{Melbourne}, \state{VIC}, \postcode{3086}, \country{Australia}}}

\affil[2]{\orgdiv{School of ICT}, \orgname{Griffith University}, \orgaddress{ \city{Gold Coast}, \state{QLD}, \postcode{4215}, \country{Australia}}}




\abstract{Anomaly detection from graph data is an important data mining task in many critical and sensitive applications such as social networks, finance, and e-commerce. Existing efforts in graph anomaly detection typically only consider the information in a single scale (view), thus inevitably limiting their capability in capturing  anomalous patterns in complex graph data. To address this limitation, in this paper, we propose a novel framework, graph \underline{\textbf{AN}}omaly d\underline{\textbf{E}}tection framework with \underline{\textbf{M}}ulti-scale c\underline{\textbf{ON}}trastive l\underline{\textbf{E}}arning (\ourmethod in short). By using a graph neural network as a backbone to encode the information from multiple graph scales (views), we learn better representation for nodes in a graph. In maximizing the agreements between instances at both the patch and context levels concurrently, we estimate the anomaly score of each node with a statistical anomaly estimator according to the degree of agreement from multiple perspectives. To further exploit a handful of ground-truth anomalies (few-shot anomalies) that may be collected in real-life applications, we further propose an extended algorithm, \ourmethodfs, to integrate valuable information in our method. 
We conduct extensive experiments under purely unsupervised settings and few-shot anomaly detection settings, and we demonstrate that our proposed method \ourmethod and its variant \ourmethodfs consistently outperforms state-of-the-art algorithms on six benchmark datasets.}




\maketitle

\section{Introduction}
\label{sec:introduction}
As a general data structure to represent  inter-dependency between objects, graphs have been widely used in many domains including social networks, biology, physics, and traffic, etc. Analyzing graph data for various tasks --- detecting anomalies from graph data in particular --- has drawn increasing attention in the research community due to its wide and critical applications in e-commence, cyber-security, and finance. For instance, by using anomaly detection algorithms in e-commerce, we can detect fraudulent sellers by jointly considering their properties and behaviors \cite{pourhabibi2020fraud}. Similarly, we can detect abnormal accounts (social bots) which spread rumors in social networks with graph anomaly detection systems \cite{latah2020detection}.

Different from conventional anomaly detection approaches for tabular/vector data where the attribute information is the only factor to be considered, graph anomaly detection requires collectively exploiting both  graph structure as well as  attribute information associated with each node. This complexity has imposed significant challenges to this task. Existing research to address the challenges can be roughly divided into two categories: (1) shallow methods, and (2) deep methods. The early shallow methods typically exploit mechanisms such as ego-network analysis \cite{amen_perozzi2016scalable}, residual analysis \cite{radar_li2017radar} or CUR decomposition \cite{anomalous_peng2018anomalous}. These methods are conceptually simple but they may be not able to learn nonlinear representation from complex graph data, leading to a sub-optimal anomaly detection performance. The deep model approaches, such as graph autoencoder (GAE) \cite{dominant_ding2019deep, li2019specae}, learn the non-linear hidden representation, and estimate the anomaly score for each node based on the reconstruction error. These methods have considerable improvements over shallow methods. However, they do not well capture the contextual information (e.g., subgraph around a node) for anomaly detection and still suffer from unsatisfactory performance. 

Very recently, a contrastive learning mechanism has been used for graph anomaly detection \cite{cola_liu2021anomaly}, which shows promising performance in graph anomaly detection. The key idea used in the proposed algorithm, namely CoLA \cite{cola_liu2021anomaly}, is to construct pairs of instances (e.g., a subgraph and a target node) and to employ a contrastive learning method to learn the representation. Based on contrastive learning, anomaly scores can be further calculated according to predictions of pairs of instances. Despite its success, only a single scale of information is considered in CoLA. In practice, due to the complexity of graph data, anomalies are often hidden in different scales (e.g., node and subgraph levels). For example, in the e-commerce application, fraudulent sellers may interact with only a small number of other users or items (i.e., local anomalies); in contrast, other cheaters may hide in larger communities (i.e., global anomalies). Such complex scales require more fine-grained and intelligent anomaly detection systems. 

Another limitation of existing graph anomaly detection methods is that these methods are designed in a purely unsupervised manner.  In scenarios where the ground-truth anomalies are unknown, these methods can make an important role. However, in practice, we may collect a handful of samples (i.e., a few shots) of anomalies. As anomalies are typically rare in many applications, these few-shot samples provide valuable information and should be incorporated into anomaly detection systems \cite{pang2019deep}. Unfortunately, most of existing graph anomaly detection methods \cite{dominant_ding2019deep, li2019specae,cola_liu2021anomaly} failed to exploit these few-shot anomalies in their design, leading to a significant information loss. 

To overcome these limitations, in this paper,  we propose a graph \underline{\textbf{AN}}omaly d\underline{\textbf{E}}tection framework with \underline{\textbf{M}}ulti-scale c\underline{\textbf{ON}}trastive l\underline{\textbf{E}}arning (\ourmethod in short) to detect anomalous nodes in graphs. Our theme is to construct multi-scales (views) from the original graphs and employ contrastive learning at both patch and context levels simultaneously to capture anomalous patterns hidden in  complex graphs. Specifically, we first employ two graph neural networks (GNNs) as encoders to learn the representation for each node and a subgraph around the target node. Then we construct pairs of positive and negative instances at both patch levels and context levels, based on which the contrastive learning maximizes the similarity between positive pairs and minimize the similarity between negative pairs. The anomaly score of each node is estimated via a novel anomaly estimator by leveraging the statistics of multi-round contrastive scores. Our framework is a general and flexible framework in the sense that it can easily incorporate ground-truth anomalies (few shot anomalies). Concretely, the labeled anomalies are seamlessly integrated into the contrastive learning framework as additional negative pairs for both patch level and context level contrastive learning, leading to a new algorithm, \ourmethodfs, for few-shot graph anomaly detection. Extensive experiments on six benchmark datasets validate the effectiveness of our algorithm \ourmethod for unsupervised graph anomaly detection and the effectiveness of \ourmethodfs for  few-shot settings. 
 The main contributions of this work are summarized as follows: 

\begin{itemize}
	\item We propose a general framework, \ourmethod, based on contrastive learning for graph anomaly detection. Our method exploits multi-scale information at both patch level and context level to capture anomalous patterns hidden in complex graphs.
	\item We present a simple approach based on the multi-scale framework, \ourmethodfs, to exploit the valuable few-shot anomalies at hand. Our method essentially enhances the flexibility of contrastive learning for anomaly detection and facilitate broader applications. 
	\item We conduct extensive experiments on six benchmark datasets to demonstrate the superiority of our \ourmethod and \ourmethodfs for both unsupervised and few-shot graph anomaly detection.
\end{itemize}

The reminder of the paper is structured as follows. We review the related works in Section \ref{sec:rw} and give the problem definition in Section \ref{sec:PD}. The proposed method \ourmethod and \ourmethodfs are described in Section \ref{sec:methodology}. The experimental results are shown in Section \ref{sec:experiments}. We conclude this paper in Section \ref{sec:conclusion}.

\section{Related Works}
\label{sec:rw}
In this section, we survey the representative works in three related topics, including graph neural networks, graph anomaly detection, and contrastive learning. 

\subsection{Graph Neural Networks}
In recent years, graph neural networks (GNNs) have achieved significant success in dealing with graph-related machine learning problems and applications \cite{gcn_kipf2017semi,gat_velivckovic2018graph,gnn_survey_wu2021comprehensive,wu2021beyond}. Considering both attributive and structural information, GNNs can learn low-dimensional representation for each node in a graph. Current GNNs can be categorized into two types: spectral and spatial methods. 
The former type of methods was initially developed on the basis of spectral theory \cite{bruna2013spectral,defferrard2016convolutional,gcn_kipf2017semi}. Bruna et al. \cite{bruna2013spectral} first extend convolution operation to graph domain using spectral graph filters. Afterward, ChebNet \cite{defferrard2016convolutional} simplifies spectral GNNs by introducing Chebyshev polynomials as the convolution filter. GCN \cite{gcn_kipf2017semi} further utilizes the first-order approximation of Chebyshev filter to learn node representations more efficiently. 
The second type of methods adopt the message-passing mechanism in the spatial domain, which propagates and aggregates local information along edges, to perform convolution operation \cite{hamilton2017inductive,gat_velivckovic2018graph,xu2019how}. GraphSAGE \cite{hamilton2017inductive} learns node representations by sampling and aggregating neighborhoods. GAT \cite{gat_velivckovic2018graph} leverages the self-attention mechanism to assign a weight for each edge when performing aggregation. GIN \cite{xu2019how} introduces a summation-based aggregation function to ensure that GNN is as powerful as the Weisfeiler-Lehman graph isomorphism test. For a thorough review, we please refer the readers to the recent survey \cite{gnn_survey_wu2021comprehensive}.

\subsection{Graph Anomaly Detection}

Anomaly detection is a conventional data mining problem aiming to identify anomalous data samples that deviate significantly from others \cite{gad_survey_ma2021comprehensive}. Compared to detecting anomalies from text/image data \cite{deepsad_ruff2019deep}, anomaly detection on graphs is often more challenging since the correlations among nodes should be also considered when measuring the abnormality of samples. To tackle the challenge, some traditional solutions use shallow mechanisms like ego-network analysis (e.g., AMEN\cite{amen_perozzi2016scalable}), residual analysis (e.g., Radar\cite{radar_li2017radar}), and CUR decomposition (e.g., ANOMALOUS \cite{anomalous_peng2018anomalous}) to model the anomalous patterns in graph data. Recently, deep learning becomes increasingly popular for graph anomaly detection \cite{gad_survey_ma2021comprehensive}. As an example of unsupervised methods, DOMINANT \cite{dominant_ding2019deep} employs a graph autoencoder model to reconstruct attribute and structural information of graphs, and the reconstruction errors are leveraged to measure node-level abnormality. CoLA \cite{cola_liu2021anomaly} considers a contrastive learning model that models abnormal patterns via learning node-subgraph agreements. Among semi-supervised methods, SemiGNN \cite{semignn_wang2019semi} is a representative method that leverages hierarchical attention to learn from multi-view graphs for fraud detection. GDN \cite{gdn_ding2021few} adopts a deviation loss to train GNN for few-shot node anomaly detection. 
Apart from the aforementioned methods for attributed graphs, some recent works also target to identity anomalies from dynamic graphs \cite{dyn_gad_wang2019detecting,taddy_liu2021anomaly}. Most recently, from the graph-level perspective, there has been research on detecting anomalies with the out-of-distribution techniques \cite{liu2024towards,shen2024optimizing, wang2024unifying}.

\subsection{Graph Contrastive Learning}

Originating from visual representation learning \cite{he2020momentum,chen2020simple,grill2020bootstrap}, contrastive learning has become increasingly popular in addressing self-supervised representation learning problems in various areas. In graph deep learning, recent works based on contrastive learning show competitive performance on graph representation learning scenario \cite{gssl_survey_liu2021graph,zheng2021towards}. DGI \cite{dgi_velickovic2019deep} learns by maximizing the mutual information between node representations and a graph-level global representation, which makes the first attempt to adapt contrastive learning in GNNs. GMI \cite{peng2020graph} jointly utilizes edge-level contrast and node-level contrast to discover high-quality node representations. 
GCC \cite{qiu2020gcc} constructs a subgraph-level contrastive learning model to learn structural representations. GCA \cite{zhu2021graph} introduces an adaptive augmentation strategy to generate different views for graph contrastive learning. MERIT \cite{Jin2021MultiScaleCS} leverages bootstrapping mechanism and multi-scale contrastiveness to learn informative node embeddings for network data.
Apart from learning effective representations, graph contrastive learning is also applied various applications, such as drug-drug interaction prediction \cite{wang2021multi} and social recommendation \cite{yu2021self}.

\section{Problem Definition}
\label{sec:PD}
In this section, we introduce and define the problem of unsupervised and few-shot graph anomaly detection. Throughout the paper, we use bold uppercase (e.g., $\mathbf{X}$), calligraphic (e.g., $\mathbfcal{V}$), and lowercase letters (e.g., $\mathbf{x}^{(i)}$) to denote matrices, sets, and vectors, respectively. We also summarize all important notations in Table \ref{table:notation}. In this work, we mainly focus on the anomaly detection tasks on attributed graphs that are widely existed in real world. Formally speaking, we define attributed graphs and graph neural networks (GNNs) as follows:

\begin{table}[!htp]
	\small
	\centering
	\caption{Summary of important notations.} 
	\begin{tabular}{ p{75 pt}<{\centering} | p{155 pt}}  
		\toprule[1.0pt]
		Symbols & Description  \\
		\cmidrule{1-2}
		$\mathcal{G}=(\mathbf{X}, \mathbf{A})$ & An attributed graph \\
		$\mathbfcal{V}, \mathbfcal{E}$ & The node and edge set of $\mathcal{G}$ \\
		$\mathbfcal{V^L}$ & The labeled node set where $\mathbfcal{V^L} \in \mathbfcal{V}$ \\
		$\mathbfcal{V^U}$ & The unlabeled node set where $\mathbfcal{V^U} \in \mathbfcal{V}$ \\
		$\mathbf{A} \in \mathbb{R}^{N \times N}$ & The adjacency matrix of $\mathcal{G}$ \\
		$\mathbf{X} \in \mathbb{R}^{N \times D}$ & The node feature matrix of $\mathcal{G}$ \\
		$\mathbf{x}^{(i)} \in \mathbb{R}^{D}$ & The feature vector of $v_i$ where $ \mathbf{x}^{(i)} \in \mathbf{X}$ \\
		$\mathbfcal{N}(v_i)$ & The neighborhood set of node $v_i \in \mathcal{V}$ \\
		\cmidrule{1-2}
		$\mathcal{G}^{(i)}_p$, $\mathcal{G}^{(i)}_c$ & Two generated subgraphs of $v_i$ \\
		$\mathbf{A}_{view}^{(i)} \in \mathbb{R}^{K \times K}$ & The adjacency matrix of $\mathcal{G}^{(i)}_{view}$ where $view \in \{p, c\}$ \\
		$\mathbf{X}_{view}^{(i)} \in \mathbb{R}^{K \times D}$ & The node feature matrix of $\mathcal{G}^{(i)}_{view}$ where $view \in \{p, c\}$ \\
		$y^{(i)}$ & The anomaly score of $v_i$ \\
		\cmidrule{1-2}
		$\mathbf{H}^{(i)}_p \in \mathbb{R}^{K \times D'}$ & The node embedding matrix of $\mathcal{G}^{(i)}_p$ \\
		$\mathbf{H}^{(i)}_c \in \mathbb{R}^{K \times D'}$ & The node embedding matrix of $\mathcal{G}^{(i)}_c$ \\
		$\mathbf{h}^{(i)}_p \in \mathbb{R}^{1 \times D'}$ & The node embeddings of masked node $v_i$ in $\mathbf{H}^{(i)}_p$ \\
		$\mathbf{h}^{(i)}_c \in \mathbb{R}^{1 \times D'}$ & The contextual embeddings of $\mathcal{G}^{(i)}_c$ \\
		$\mathbf{z}^{(i)}_p, \mathbf{z}^{(i)}_c \in \mathbb{R}^{1 \times D'}$ & The node embeddings of $v_i$ in patch-level and context-level networks  \\
		$s^{(i)}_p, \tilde{s}^{(i)}_p \in \mathbb{R}$ & The positive and negative patch-level contrastive scores of $v_i$ \\
		$s^{(i)}_c, \tilde{s}^{(i)}_c \in \mathbb{R}$ & The positive and negative context-level contrastive scores of $v_i$ \\
		$\mathbf{\Theta}, \mathbf{\Phi} \in \mathbb{R}^{D \times D'}$ & The trainable parameter matrices of two graph encoders \\
		$\mathbf{W}_{p},\mathbf{W}_{c}  \in \mathbb{R}^{D' \times D'}$ & The trainable parameter matrices of two bilinear mappings \\
		\cmidrule{1-2}
		$N^L, N^U, N$ & The number of labeled, unlabeled, and all nodes in $\mathcal{G}$ \\
		$K$ & The number of nodes in subgraphs \\
		$D$ & The dimension of node attributes in $\mathcal{G}$ \\
 		$D'$ & The dimension of node embeddings  \\
		$R$ & The number of evaluation rounds in anomaly scoring \\
		\bottomrule[1.0pt]
	\end{tabular}
	\label{table:notation}
\end{table}

\begin{definition}[Attributed Graphs]
Given an attributed graph $\mathcal{G}=(\mathbf{X},\mathbf{A})$, we denote its node attribute (i.e., feature) and adjacency matrices as $\mathbf{X} \in \mathbb{R}^{N \times D}$ and $\mathbf{A} \in \mathbb{R}^{N \times N}$, where $N$ and $D$ are the number of nodes and feature dimensions. 
An attribute graph can also be defined as $\mathcal{G}=(\mathbfcal{V}, \mathbfcal{E}, \mathbf{X})$, where $\mathbfcal{V}=\{v_1, v_2, \cdots, v_N\}$ and $\mathbfcal{E}=\{e_1, e_2, \cdots, e_M\}$ are node and edge sets. Thus, we have $N=|\mathbfcal{V}|$, the number of edges $M=|\mathbfcal{E}|$, and define $\mathbf{x}_i \in \mathbb{R}^{D}$ as the attributes of node $v_i$. 
To represent the underlying node connectivity, we let $\mathbf{A}_{ij}=1$ if there exists an edge between $v_i$ and $v_j$ in $\mathbfcal{E}$, otherwise $\mathbf{A}_{ij}=0$. 
In particular, given a node $v_i$, we define its neighborhood set as $\mathbfcal{N}(v_i)=\{v_j \in \mathbfcal{V} |  \mathbf{A}_{ij} \neq 0 \}$.
\end{definition}

\begin{definition}[Graph Neural Networks]
Given an attributed graph $\mathcal{G}=(\mathbf{X},\mathbf{A})$, a parameterized graph neural network $GNN(\cdot)$ aims to learn the low-dimensional embeddings of $\mathbf{X}$ by considering the topological information $\mathbf{A}$, denoted as $\mathbf{H} \in \mathbb{R}^{N \times D'}$, where $D'$ is the embedding dimensions and we have $D' \ll D$. For a specific node $v_i \in \mathbfcal{V}$, we denote its embedding as $\mathbf{h}^{(i)} \in \mathbb{R}^{D'}$ where $\mathbf{h}^{(i)} \in \mathbf{H}$.
\end{definition}

In this paper, we focus on two different anomaly detection tasks on attributed graphs, namely unsupervised and few-shot graph anomaly detection. 
Firstly, we define the problem of unsupervised graph anomaly detection as follows:

\begin{definition}[Unsupervised Graph Anomaly Detection]
Given an unlabeled attribute graph $\mathcal{G}=(\mathbf{X},\mathbf{A})$, we intend to train and evaluate a graph anomaly detection model $\mathcal{F}(\cdot): \mathbb{R}^{N \times D} \to \mathbb{R}^{N \times 1}$ across all nodes in $\mathbfcal{V}$, where we use $\mathbf{y}$ to denote the output node anomaly scores, and $y^{(i)}$ is the anomaly score of node $v_i$.
\end{definition}

For graphs with limited prior knowledge (i.e., labeling information) on the underlying anomalies, we define the problem of few-shot graph anomaly detection as below:

\begin{definition}[Few-shot Graph Anomaly Detection]
For an attribute graph $\mathcal{G}=(\mathbf{X},\mathbf{A})$, we have a small set of labeled anomalies $\mathbfcal{V}^L$ and the rest set of unlabeled nodes $\mathbfcal{V}^U$, where $|\mathbfcal{V}^L| \ll |\mathbfcal{V}^U|$ since it is relatively expensive to label anomalies in the real world so that only a very few labeled anomalies are typically available. Thus, our goal is to learn a model $\mathcal{F}(\cdot): \mathbb{R}^{N \times D} \to \mathbb{R}^{N \times 1}$ on $\mathbfcal{V}^L \cup \mathbfcal{V}^U$, which measures node abnormalities by calculating their anomaly scores $\mathbf{y}$. It is worth noting that during the evaluation, the well-trained model $\mathcal{F}^*(\cdot)$ is only tested on $\mathbfcal{V}^U$ to prevent the potential information leakage.
\end{definition}



\section{Methodology}
\label{sec:methodology}
\begin{figure*}[t]
\centering
\includegraphics[width=1.0\textwidth]{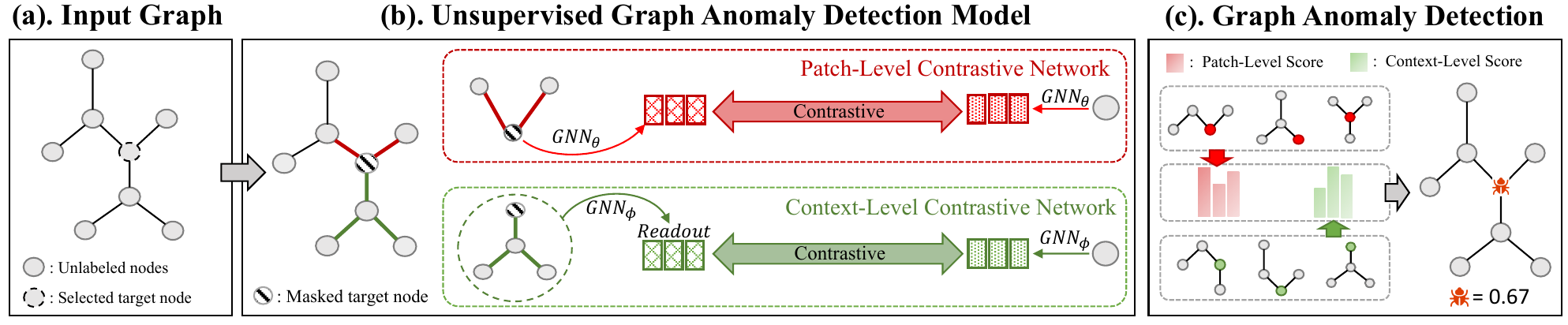}
\caption{
The conceptual framework of \ourmethod. Given an attributed graph $\mathcal{G}$, we first sample a batch of target nodes, where their associated anonymized subgraphs are generated and fed into two contrastive networks. Then, we design a multi-scale (i.e., patch-level and context-level) contrastive network to learn agreements between node and contextual embeddings from different perspectives. During the model inference, two contrastive scores are statistically annealed to obtain the final anomaly score of each node in $\mathcal{G}$.
}
\label{fig:framework1}
\end{figure*}

In this section, we introduce the proposed \ourmethod and \ourmethodfs algorithms in detecting node-level graph anomalies in an unsupervised and few-shot supervised manner. The overall frameworks of our methods are shown in Figure \ref{fig:framework1} and \ref{fig:framework2}, which consist of four main components, namely the \textit{augmented subgraphs generation}, \textit{patch-level contrastive network}, \textit{context-level contrastive network}, and \textit{statistical graph anomaly scorer}. Firstly, given a target node from the input graph, we exploit its contextual information by generating two subgraphs associated with it. Then, we propose two general yet powerful contrastive mechanisms for graph anomaly detection tasks. Specifically, for an attributed graph without any prior knowledge on the underlying anomalies, the proposed \ourmethod method learns the patch-level and context-level agreements by maximizing (1) the mutual information between node embeddings in the patch-level contrastive network and (2) the mutual information between node embeddings and their contextual embeddings in the context-level contrastive network. The underlying intuition is that there are only a few anomalies, and thus our well-trained model can identify the salient attributive and structural mismatch between an abnormal node and its surrounding contexts by throwing a significantly higher contrastive score. On the other hand, if few labeled anomalies are available in an attributed graph, the proposed \ourmethodfs variant can effectively utilize the limited labeling information to further enrich the supervision signals extracted by \ourmethod. This intriguing capability is achieved by plugging in a different contrastive route, where the aforementioned patch-level and context-level agreements are minimized for labeled anomalies while still maximized for unlabeled nodes as same as in \ourmethod. 
Finally, we design a universal graph anomaly scorer to measure node abnormalities by statistically annealing the patch-level and context-level contrastive scores at the inference stage, which shares and works on both unsupervised and few-shot supervised scenarios.

In the rest of this section, we introduce the four primary components of \ourmethod in Subsection \ref{subsec: augmented subg generation}, \ref{subsec: patch-level}, \ref{subsec: context-level}, and \ref{subsec: scorer}. Particularly, in Subsection \ref{subsec: few-shot}, we discuss how \ourmethod can be extended to more competitive \ourmethodfs in detail to incorporate the available supervision signals provided by a few labeled anomalies. In Subsection \ref{subsec: optimization}, we present and discuss the training objective of \ourmethod and \ourmethodfs, as well as their algorithms and time complexity.

\subsection{Augmented Subgraphs Generation} \label{subsec: augmented subg generation}
\begin{figure*}[t]
\centering
\includegraphics[width=1.0\textwidth]{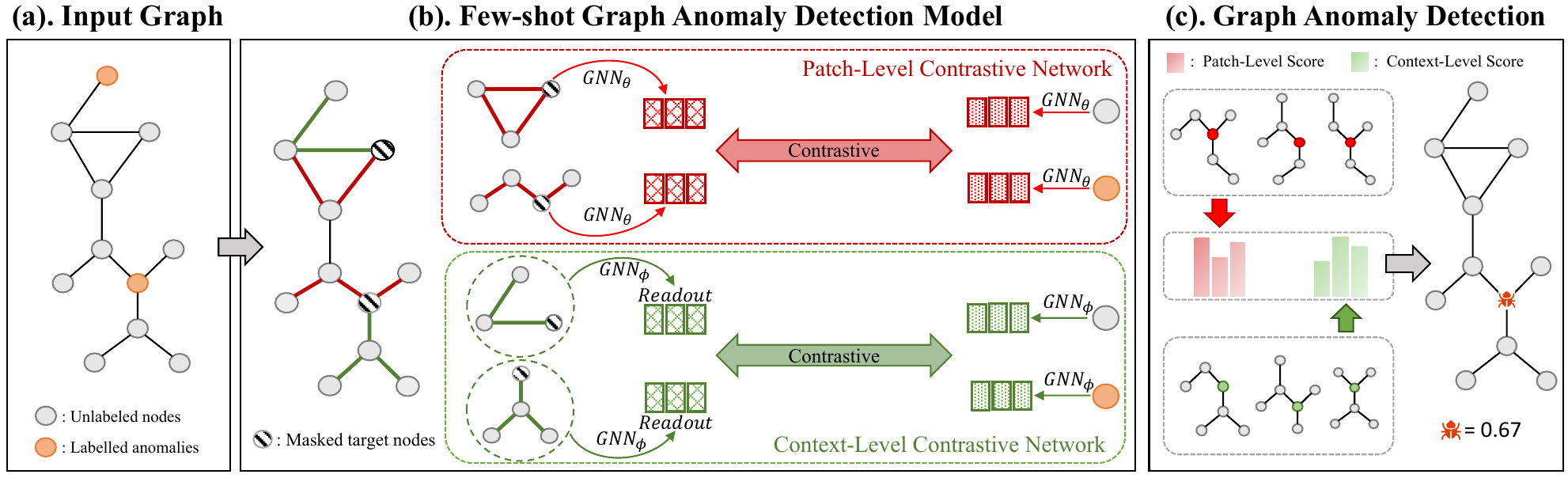}
\caption{
The conceptual framework of \ourmethodfs, which shares the similar pipeline of \ourmethod. Given an attributed graph $\mathcal{G}$, we first sample a batch of target nodes. After this, we design a different multi-scale contrastive network equipped with two contrastive routes, where the agreements between node and contextual embeddings are maximized for unlabeled node while minimized for labeled anomalies in a mini-batch. At the inference stage, it has the same graph anomaly detector to estimate the anomlay socre of each node in $\mathcal{G}$.}
\label{fig:framework2}
\end{figure*}

Graph contrastive learning relies on effective discrimination pairs to extract supervision signals from the rich graph attributive and topological information \cite{gssl_survey_liu2021graph}. Recently, graph augmentations, such as attribute masking, edge modification, subgraph sampling, and graph diffusion, are widely applied to assist contrastive models in learning expressive graph representations \cite{Jin2021MultiScaleCS, zheng2021towards}. However, not all of them are directly applicable to anomaly detection tasks. For example, edge modification and graph diffusion can distort the original topological information, thus hindering the model to distinguish an abnormal node from its surrounding contexts effectively. 
To avoid falling into this trap, we adopt an \textit{anonymized subgraph sampling} mechanism to generate graph views for our contrastive networks, which based on two motivations: (1) The agreement between a node and its surrounding contexts (i.e., subgraphs) is typically sufficient to reflect the abnormality of this node \cite{jin2021anemone, zheng2021generative}; (2) Subgraph sampling provides adequate diversity of node surrounding contexts for robust model training and statistical anomaly scoring. Specifically, we explain the details of the proposed augmentation strategy for graph anomaly detection as follows:

\begin{enumerate}
    \item \textbf{Target node sampling.} 
    As this work mainly focuses on node-level anomaly detection, we first sample a batch of target nodes from a given attributed graph. It is worth noting that for a specific target node, it may associate with the label or not, which results in different contrastive routes as shown in the middle red and green dashed boxes in Figures \ref{fig:framework1} and \ref{fig:framework2}. We discuss this in detail in the following subsections.
    
    \item \textbf{Surrounding context sampling.} 
    Although several widely adopted graph augmentations are available \cite{gssl_survey_liu2021graph}, most of them are designed to slightly attack the original graph attributive or topological information to learn robust and expressive node-level or graph-level representations, which violate our motivations mentioned above and introduce extra anomalies. Thus, in this work, we employ the subgraph sampling as the primary augmentation strategy to generate augmented graph views for each target nodes based on the random walk with restart (RWR) algorithm \cite{tong2006fast}. Taking a target node $v_i$ for example, we generate its surrounding contexts by sampling subgraphs centred at it with a fixed size $K$, denoted as $\mathcal{G}^{(i)}_p = (\mathbf{A}^{(i)}_p, \mathbf{X}^{(i)}_p)$ and $\mathcal{G}^{(i)}_c = (\mathbf{A}^{(i)}_c, \mathbf{X}^{(i)}_c)$ for patch-level and context-level contrastive networks. In particular, we let the first node in $\mathcal{G}^{(i)}_p$ and $\mathcal{G}^{(i)}_c$ as the starting (i.e., target) node.
    
    \item \textbf{Target node anonymization.} Although the above-generated graph views can be directly fed into the contrastive networks, there is a critical limitation: The attributive information of target nodes involves calculating their patch-level and context-level embeddings, which results in information leakage during the multi-level contrastive learning. To prevent this issue and construct harder pretext tasks to boost model training \cite{gssl_survey_liu2021graph}, we anonymize target nodes in their graph views by completely masking their attributes, i.e., $\mathbf{X}^{(i)}_p[1,:] \rightarrow \overrightarrow{0}$ and $\mathbf{X}^{(i)}_c[1,:] \rightarrow \overrightarrow{0}$.
\end{enumerate}

\subsection{Patch-level Contrastive Network} \label{subsec: patch-level}
The objective of patch-level contrastiveness is to learn the local agreement between the embedding of masked target node $v_i$ in its surrounding contexts $\mathcal{G}^{(i)}_p$ and the embedding of $v_i$ itself. The underlying intuition of the proposed patch-level contrastive learning is that the mismatch between a node and its directly connected neighbors is an effective measurement to detect \textit{local anomalies}, which indicates the anomalies that are distinguishable from their neighbors. For example, some e-commerce fraudsters are likely to transact with unrelated users directly, where patch-level contrastiveness is proposed to detect such anomalies in an attributed graph.
As shown in Figure \ref{fig:framework1}, our patch-level contrastive network consists of two main components: Graph encoder and contrastive module. 

\vspace{2mm}\noindent \textbf{Graph encoder.} The patch-level graph encoder takes a target node and one of its subgraph (i.e., surrounding context $\mathcal{G}^{(i)}_p$) as the input. Specifically, the node embeddings of $\mathcal{G}^{(i)}_p$ are calculated in below:

\begin{equation}
\begin{aligned}
\mathbf{H}^{(i)}_{p}  &= GNN_{\theta}\left(\mathcal{G}^{(i)}_p\right) = GCN\left(\mathbf{A}^{(i)}_p, \mathbf{X}^{(i)}_p ; \mathbf{\Theta}  \right)\\
 &= \sigma\left(\widetilde{{\mathbf{D}}^{(i)}_p}^{-\frac{1}{2}} \widetilde{{\mathbf{A}}^{(i)}_p} \widetilde{{\mathbf{D}}^{(i)}_p}^{-\frac{1}{2}} \mathbf{X}^{(i)}_p \mathbf{\Theta} \right),
\end{aligned}
\label{eq:gnn}
\end{equation}
where $\mathbf{\Theta} \in \mathbb{R}^{D \times D'}$ denotes the set of trainable parameters of patch-level graph neural network $GNN_{\theta}(\cdot)$. For simplicity and follow \cite{jin2021anemone}, we adopt a single layer graph convolution network (GCN) \cite{gcn_kipf2017semi} as the backbone encoder, where $\sigma(\cdot)$ denotes the ReLU activation in a typical GCN layer, $\widetilde{{\mathbf{A}}^{(i)}_p} = \mathbf{A}^{(i)}_p + \mathbf{I}$, and $\widetilde{{\mathbf{D}}^{(i)}_p}$ is the calculated degree matrix of $\mathcal{G}^{(i)}_p$ by row-wise summing $\widetilde{{\mathbf{A}}^{(i)}_p}$. Alternatively, one may also replace GCN with other off-the-shelf graph neural networks to aggregate messages from nodes' neighbors to calculate $\mathbf{H}^{(i)}_{p}$.

In patch-level contrastiveness, our discrimination pairs are the masked and original target node embeddings (e.g., $\mathbf{h}^{(i)}_{p}$ and $\mathbf{z}^{(i)}_{p}$ for a target node $v_i$), where the former one can be easily obtained via $\mathbf{h}^{(i)}_{p} = \mathbf{H}^{(i)}_{p}[1,:]$. To calculate the embeddings of original target nodes, e.g., $\mathbf{z}^{(i)}_{p}$, we only have to fed $\mathbf{x}^{(i)} = \mathbf{X}[i,:]$ into $GNN_{\theta}(\cdot)$ without the underlying graph structure since there is only a single node $v_i$. In such a way, $GNN_{\theta}(\cdot)$ degrades to a MLP that is parameterized with $\mathbf{\Theta}$. We illustrate the calculation of $\mathbf{z}^{(i)}_{p}$ as follows:

\begin{equation}
\mathbf{z}^{(i)}_{p}  = GNN_{\theta}\left(\mathbf{x}^{(i)}\right) = \sigma\left(\mathbf{x}^{(i)} \mathbf{\Theta} \right),
\label{eq:mlp}
\end{equation}
where adopting $\mathbf{\Theta}$ ensures $\mathbf{z}^{(i)}_{p}$ and $\mathbf{h}^{(i)}_{p}$ are mapped into the same latent space to assist the following contrasting. 

\vspace{2mm}\noindent \textbf{Patch-level Contrasting.} To measure the agreement between $\mathbf{h}^{(i)}_{p}$ and $\mathbf{z}^{(i)}_{p}$, we adopt a bilinear mapping to compute the similarity between them (i.e., the positive score in \ourmethod), denoted as $\mathbf{s}^{(i)}_{p}$:

\begin{equation}
\mathbf{s}^{(i)}_{p}  = Bilinear\left( \mathbf{h}^{(i)}_{p}, \mathbf{z}^{(i)}_{p} \right) = \sigma\left(\mathbf{h}^{(i)}_{p} \mathbf{W}_p {\mathbf{z}^{(i)}_{p}}^\top \right),
\label{eq:patch-level anemone positive score}
\end{equation}
where $W_p \in \mathbb{R}^{D' \times D'}$ is a set of trainable weighting parameters, and $\sigma(\cdot)$ denotes the Sigmoid activation in this equation.

Also, there is a \textit{patch-level negative sampling} mechanism to assist model training and avoid it being biased by merely optimizing on positive pairs. Specifically, we first calculate $\mathbf{h}^{(j)}_{p}$ based on the subgraph centred at an irrelevant node $v_j$, then we calculate the similarity between $\mathbf{h}^{(j)}_{p}$ and $\mathbf{z}^{(i)}_{p}$ (i.e., negative score) with the identical bilinear mapping:

\begin{equation}
\tilde{\mathbf{s}}^{(i)}_{p}  = Bilinear\left( \mathbf{h}^{(j)}_{p}, \mathbf{z}^{(i)}_{p} \right) = \sigma\left(\mathbf{h}^{(j)}_{p} \mathbf{W}_p {\mathbf{z}^{(i)}_{p}}^\top \right).
\label{eq:patch-level anemone negative score}
\end{equation}

In practice, we train the model in a mini-batch manner as mentioned in Subsection \ref{subsec: augmented subg generation}. Thus, $\mathbf{h}^{(j)}_{p}$ can be easily acquired by using other masked target node embeddings in the same mini-batch with size $B$. Finally, the patch-level contrastive objective of \ourmethod (under the context of unsupervised graph anomaly detection) can be formalized with the Jensen-Shannon divergence \cite{dgi_velickovic2019deep}:

\begin{equation}
\mathcal{L}_{p}=-\frac{1}{2n}\sum_{i=1}^{B}\left(log\left(\mathbf{s}^{(i)}_{p}\right)+log\left(1-\tilde{\mathbf{s}}^{(i)}_{p}\right)\right).
\label{eq:patch-level loss}
\end{equation}

\subsection{Context-level Contrastive Network} \label{subsec: context-level}
Different from the patch-level contrastiveness, the objective of context-level contrasting is to learn the global agreement between the contextual embedding of a masked target node $v_i$ in $\mathcal{G}^{(i)}_c$ and the embedding of itself by mapping its attributes to the latent space. The intuition behind this is to capture the \textit{global anomalies} that are difficult to be distinguished by directly comparing with the closest neighbors. For instance, the fraudsters are also likely to camouflage themselves in large communities, resulting in a more challenging anomaly detection task. To enable the model to detect these anomalies, we propose the context-level contrastiveness with a multi-scale (i.e., node versus graph) contrastive learning schema. In the middle part of Figure \ref{fig:framework1}, we illustrate the conceptual design of our context-level contrastive network, which has three main components: graph encoder, readout module, and contrastive module. 

\vspace{2mm}\noindent \textbf{Graph encoder and readout module.} The context-level graph encoder shares the identical neural architecture of the patch-level graph encoder, but it has a different set of trainable parameters $\mathbf{\Phi}$. Specifically, given a subgraph $\mathcal{G}^{(i)}_c$ centred at the target node $v_i$,  we calculate the node embeddings of $\mathcal{G}^{(i)}_c$ in a similar way:

\begin{equation}
\mathbf{H}^{(i)}_{c}  = GNN_{\phi}\left(\mathcal{G}^{(i)}_c\right) = \sigma\left(\widetilde{{\mathbf{D}}^{(i)}_c}^{-\frac{1}{2}} \widetilde{{\mathbf{A}}^{(i)}_c} \widetilde{{\mathbf{D}}^{(i)}_c}^{-\frac{1}{2}} \mathbf{X}^{(i)}_c \mathbf{\Phi} \right).
\label{eq:gnn2}
\end{equation}

The main difference between the context-level and patch-level contrastiveness is that the former aims to contrast target node embeddings with subgraph embeddings (i.e., node versus subgraph), while the aforementioned patch-level contrasting learns the agreements between the masked and original target node embeddings (i.e., node versus node).
To obtain the contextual embedding of target node $v_i$ (i.e., $\mathbf{h}^{(i)}_{c}$), we aggregate all node embeddings in $\mathbf{H}^{(i)}_{c}$ with an average readout function:

\begin{equation}
\mathbf{h}^{(i)}_{c} = readout \left( \mathbf{H}^{(i)}_{c} \right) = \frac{1}{K}\sum_{j=1}^{K}\mathbf{H}^{(i)}_{c}[j,:],
\label{eq:readout}
\end{equation}
where $K$ denotes the number of nodes in a contextual subgraph.

Similarly, we can also obtain the embedding of $v_i$ via a non-linear mapping:

\begin{equation}
\mathbf{z}^{(i)}_{c}  = GNN_{\phi}\left(\mathbf{x}^{(i)}\right) = \sigma\left(\mathbf{x}^{(i)} \mathbf{\Phi} \right),
\label{eq:mlp2}
\end{equation}
where $\mathbf{z}^{(i)}_{c}$ and $\mathbf{h}^{(i)}_{c}$ are projected to the same latent space with a shared set of parameters $\mathbf{\Phi}$. 

\vspace{2mm}\noindent \textbf{Context-level contrasting.} We measure the similarity between $\mathbf{h}^{(i)}_{c}$ and $\mathbf{z}^{(i)}_{c}$ (i.e., the positive score in \ourmethod) with a different parameterized bilinear function:

\begin{equation}
\mathbf{s}^{(i)}_{c}  = Bilinear\left( \mathbf{h}^{(i)}_{c}, \mathbf{z}^{(i)}_{c} \right) = \sigma\left(\mathbf{h}^{(i)}_{c} \mathbf{W}_c {\mathbf{z}^{(i)}_{c}}^\top \right),
\label{eq:context-level anemone positive score}
\end{equation}
where $W_c \in \mathbb{R}^{D' \times D'}$ and $\sigma(\cdot)$ is the Sigmoid activation. Similarly, there is a \textit{context-level negative sampling} mechanism to avoid model collapse, where negatives $\mathbf{h}^{(j)}_{c}$ are obtained from other irrelevant surrounding contexts $\mathcal{G}^{(j)}_c$ where $j \neq i$. Thus, the negative score can be obtained via:

\begin{equation}
\tilde{\mathbf{s}}^{(i)}_{c}  = Bilinear\left( \mathbf{h}^{(j)}_{c}, \mathbf{z}^{(i)}_{c} \right) = \sigma\left(\mathbf{h}^{(j)}_{c} \mathbf{W}_c {\mathbf{z}^{(i)}_{c}}^\top \right).
\label{eq:context-level anemone negative score}
\end{equation}

Finally, the context-level contrastiveness is ensured by optimizing the following objective:

\begin{equation}
\mathcal{L}_{c}=-\frac{1}{2n}\sum_{i=1}^{B}\left(log\left(\mathbf{s}^{(i)}_{c}\right)+log\left(1-\tilde{\mathbf{s}}^{(i)}_{c}\right)\right).
\label{eq:context-level loss}
\end{equation}

\subsection{Few-shot Multi-scale Contrastive Network} \label{subsec: few-shot}
The above patch-level and context-level contrastiveness are conceptually designed to detect attributive and structural graph anomalies in an unsupervised manner (i.e., the proposed \ourmethod method in Algorithm \ref{algo: anemone}). However, how to incorporate limited supervision signals remains unknown if there are a few available labeled anomalies. To answer this question, we proposed an extension of \ourmethod named \ourmethodfs to perform graph anomaly detection in a few-shot supervised manner without drastically changing the overall framework (i.e., Figure \ref{fig:framework2}) and training objective (i.e., Equation \ref{eq:patch-level loss}, \ref{eq:context-level loss}, and \ref{eq:loss}). This design further boosts the performance of \ourmethod significantly with only a few labeled anomalies, which can be easily acquired in many real-world applications. Specifically, given a mini-batch of target nodes $\mathbfcal{V}_B=\{\mathbfcal{V}^L_B, \mathbfcal{V}^U_B\}$ with labeled anomalies and unlabeled nodes, we design and insert a different contrastive route in the above patch-level and context-level contrastiveness, as the bottom dashed arrows, i.e., the so-called negative pairs, in the red and green dashed boxes shown in the middle part of Figure \ref{fig:framework2}. 

\vspace{2mm}\noindent \textbf{Few-shot patch-level contrasting.} For nodes in $\mathbfcal{V}^U_B$, we follow Equation \ref{eq:patch-level anemone positive score} and \ref{eq:patch-level anemone negative score} to compute positive and negative scores as in \ourmethod. However, for a target node $v_k$ in $\mathbfcal{V}^L_B$, we minimize the mutual information between its masked node embedding $\mathbf{h}^{(k)}_{p}$ and original node embedding $\mathbf{z}^{(k)}_{p}$, which equivalents to enrich the patch-level negative set with an additional negative pair:

\begin{equation}
\tilde{\mathbf{s}}^{(k)}_{p}  = Bilinear\left( \mathbf{h}^{(k)}_{p}, \mathbf{z}^{(k)}_{p} \right) = \sigma\left(\mathbf{h}^{(k)}_{p} \mathbf{W}_p {\mathbf{z}^{(k)}_{p}}^\top \right).
\label{eq:patch-level anemone-fs negative score}
\end{equation}

The behind intuitions are in two-folds. Firstly, for most unlabeled nodes, we assume that most of them are not anomalies so that the mutual information between masked and original target node embeddings should be maximized for the model to distinguish normal nodes from a few anomalies in an attributed graph. Secondly, for a labeled target node (i.e., an anomaly), this mutual information should be minimized for the model to learn how anomalies should be different from their surrounding contexts. As a result, there are two types of patch-level negative pairs in \ourmethodfs: (1) $\mathbf{h}^{(j)}_{p}$ and $\mathbf{z}^{(i)}_{p}$ where $v_i \in \mathbfcal{V}^U_B$ and $i \neq j$; (2) $\mathbf{h}^{(k)}_{p}$ and $\mathbf{z}^{(k)}_{p}$ where $v_k \in \mathbfcal{V}^L_B$. 

\vspace{2mm}\noindent \textbf{Few-shot context-level contrasting.} Similarly, for a node $v_k$ in $\mathbfcal{V}^L_B$, we minimize the mutual information between its contextual embedding $\mathbf{h}^{(k)}_{c}$ and the embedding of itself $\mathbf{h}^{(k)}_{c}$ by treating them as a negative pair:

\begin{equation}
\tilde{\mathbf{s}}^{(k)}_{c}  = Bilinear\left( \mathbf{h}^{(k)}_{c}, \mathbf{z}^{(k)}_{c} \right) = \sigma\left(\mathbf{h}^{(k)}_{c} \mathbf{W}_c {\mathbf{z}^{(k)}_{c}}^\top \right).
\label{eq:context-level anemone-fs negative score}
\end{equation}

The behind intuition is the same as in few-shot patch-level contrasting, and we also have two types of negatives in this module to further assist the model in obtaining richer supervision signals from various perspectives. \\

In general, this proposed plug-and-play extension enhances the original self-supervise contrastive anomaly detection mechanism in \ourmethod by constructing extra negative pairs with the available limited labeled anomalies, enabling the model to achieve better performance in more realistic application scenarios.

\subsection{Statistical Graph Anomaly Scorer} \label{subsec: scorer}
So far, we have introduced two contrastive mechanisms in \ourmethod and \ourmethodfs for different graph anomaly detection tasks. After the model is well-trained, we propose an universal statistical graph anomaly scorer for both \ourmethod and \ourmethodfs to calculate the anomaly score of each node in $\mathcal{G}$ during the model inference. Specifically, we first generate $R$ subgraphs centred at a target node $v_i$ for both contrastive networks. Then, we calculate patch-level and context-level contrastive scores accordingly, i.e., $[\mathbf{s}^{(i)}_{p,1}, \cdots, \mathbf{s}^{(i)}_{p,R}, \mathbf{s}^{(i)}_{c,1}, \cdots, \mathbf{s}^{(i)}_{c,R},\tilde{\mathbf{s}}^{(i)}_{p,1}, \cdots, \tilde{\mathbf{s}}^{(i)}_{p,R}, \tilde{\mathbf{s}}^{(i)}_{c,1}, \cdots, \tilde{\mathbf{s}}^{(i)}_{c,R}]$. After this, we define the base patch-level and context-level anomaly scores of $v_i$ as follows:

\begin{equation}
b^{(i)}_{view,j} = \tilde{\mathbf{s}}^{(i)}_{view,j} - \mathbf{s}^{(i)}_{view,j}, 
\label{eq:base score}
\end{equation}
where $j \in \{1, \cdots, R\}$, and ``view" corresponds to $p$ or $c$ to denote the patch-level or context-level base score, respectively. If $v_i$ is a normal node, then $\mathbf{s}^{(i)}_{view,j}$ and $\tilde{\mathbf{s}}^{(i)}_{view,j}$ are expected to be close to 1 and 0, leading $b^{(i)}_{view,j}$ close to -1. Otherwise, if $v_i$ is an anomaly, then $\mathbf{s}^{(i)}_{view,j}$ and $\tilde{\mathbf{s}}^{(i)}_{view,j}$ are close to 0.5 due to the mismatch between $v_i$ and its surrounding contexts, resulting in $b^{(i)}_{view,j} \rightarrow 0$. Therefore, we have $b^{(i)}_{view,j}$ in the range of $[-1, 0]$.

Although we can directly use the base anomaly score $b^{(i)}_{view,j}$ to indicate whether $v_i$ is an anomaly, we design a more sophisticated statistical abnormality scorer to calculate the final patch-level and context-level anomaly scores $y^{(i)}_{view}$ of $v_i$ based on $b^{(i)}_{view,j}$:

\begin{equation}
\begin{aligned}
\bar{b}^{(i)}_{view} &={\sum_{j=1}^{R} b^{(i)}_{view,j} }/{R}, \\
y^{(i)}_{view} &= \bar{b}^{(i)}_{view} + \sqrt{{\sum_{j=1}^{R}\left(b^{(i)}_{view,j} - \bar{b}^{(i)}_{view}\right)^{2}}/{R}}.
\end{aligned}
\label{eq:final score v1}
\end{equation}

The underlying intuitions behind the above equation are: (1) An abnormal node usually has a larger base anomaly score; (2) The base scores of an abnormal node are typically unstable (i.e., with a larger standard deviation) under $R$ evaluation rounds. 
Finally, we anneal $y^{(i)}_{p}$ and $y^{(i)}_{c}$ to obtain the final anomaly score of $v_i$ with a tunable hyper-parameter $\alpha \in [0, 1]$ to balance the importance of two anomaly scores at different scales:

\begin{equation}
y^{(i)} = \alpha y^{(i)}_{c} + (1 - \alpha) y^{(i)}_{p}.
\label{eq:final score v2}
\end{equation}

\begin{algorithm}[t]
	\caption{The Proposed \ourmethod Algorithm}
	\label{algo: anemone}
    \textbf{Input}: Attributed graph $\mathcal{G}$ with a set of unlabeled nodes $\mathbfcal{V}$; Maximum training epochs $E$; Batch size $B$; Number of evaluation rounds $R$. \\
    \textbf{Output}: Well-trained graph anomaly detection model $\mathcal{F}^{*}(\cdot)$. \\ \vspace{-4mm}
    \begin{algorithmic}[1]
		\STATE Randomly initialize the trainable parameters $\mathbf{\Theta}$, $\mathbf{\Phi}$, $\mathbf{W}_p$, and $\mathbf{W}_c$;
		\STATE $/*$ {\it Model training} $*/$
		\FOR{$e \in 1,2,\cdots,E$}
			\STATE $\mathbfcal{B} \leftarrow$ Randomly split $\mathbfcal{V}$ into batches with size $B$;
			\FOR{batch $\widetilde{\mathbfcal{B}}=(v_{1},\cdots,v_{B}) \in \mathbfcal{B}$}
			\STATE Sample two anonymized subgraphs for each node in $\widetilde{\mathbfcal{B}}$, i.e., $\{\mathcal{G}_p^{(1)},\cdots,\mathcal{G}_p^{(B)}\}$ and $\{\mathcal{G}_c^{(1)},\cdots,\mathcal{G}_c^{(B)}\}$;
			\STATE Calculate the masked and original node embeddings via Eq. \eqref{eq:gnn}, \eqref{eq:mlp};
	        \STATE Calculate the masked node and its contextual embeddings via Eq. \eqref{eq:gnn2}, \eqref{eq:readout}, and \eqref{eq:mlp2};
            \STATE Calculate the patch-level positive and negative scores for for each node in $\widetilde{\mathbfcal{B}}$ via Eq. \eqref{eq:patch-level anemone positive score} and \eqref{eq:patch-level anemone negative score};
            \STATE Calculate the context-level positive and negative scores for for each node in $\widetilde{\mathbfcal{B}}$ via Eq. \eqref{eq:context-level anemone positive score} and \eqref{eq:context-level anemone negative score};
            \STATE Calculate the loss $\mathcal{L}$ via Eq. \eqref{eq:patch-level loss}, \eqref{eq:context-level loss}, and \eqref{eq:loss};
			\STATE Back propagate to update trainable parameters $\mathbf{\Theta}$, $\mathbf{\Phi}$, $\mathbf{W}_p$, and $\mathbf{W}_c$;
			\ENDFOR
		\ENDFOR
		\STATE $/*$ {\it Model inference} $*/$
		\FOR{$v_i \in \mathcal{V}$}
		\FOR{evaluation round $r \in 1,2,\cdots,R$}
		\STATE Calculate $b^{(i)}_p$ and $b^{(i)}_c$ via Eq. \eqref{eq:base score};
		\ENDFOR
		\STATE Calculate the final patch-level and context-level anomaly scores $y^{(i)}_p$ and $y^{(i)}_c$ over $R$ evaluation rounds via Eq. \eqref{eq:final score v1};
		\STATE Calculate the final anomaly score $y^{(i)}$ of $v_i$ via Eq. \eqref{eq:final score v2};
		\ENDFOR
	\end{algorithmic}
\end{algorithm}


\begin{algorithm}[t]
	\caption{The Proposed \ourmethodfs Algorithm}
	\label{algo: anemone-fs}
    \textbf{Input}: Attributed graph $\mathcal{G}$ with a set of labeled and unlabeled nodes $\mathbfcal{V}=\{\mathbfcal{V}^L, \mathbfcal{V}^U\}$ where $|\mathbfcal{V}^L| \ll |\mathbfcal{V}^U|$; Maximum training epochs $E$; Batch size $B$; Number of evaluation rounds $R$. \\
    \textbf{Output}: Well-trained graph anomaly detection model $\mathcal{F}^{*}(\cdot)$. \\ \vspace{-4mm}
    \begin{algorithmic}[1]
		\STATE Randomly initialize the trainable parameters $\mathbf{\Theta}$, $\mathbf{\Phi}$, $\mathbf{W}_p$, and $\mathbf{W}_c$;
		\STATE $/*$ {\it Model training} $*/$
		\FOR{$e \in 1,2,\cdots,E$}
			\STATE $\mathbfcal{B} \leftarrow$ Randomly split $\mathbfcal{V}$ into batches with size $B$;
			\FOR{batch $\widetilde{\mathbfcal{B}}=\{\mathbfcal{V}^L_B, \mathbfcal{V}^U_B\}=(v_{1},\cdots,v_{B}) \in \mathbfcal{B}$}
			\STATE Sample two anonymized subgraphs for each node in $\widetilde{\mathbfcal{B}}$, i.e., $\{\mathcal{G}_p^{(1)},\cdots,\mathcal{G}_p^{(B)}\}$ and $\{\mathcal{G}_c^{(1)},\cdots,\mathcal{G}_c^{(B)}\}$;
			\STATE Calculate the masked and original node embeddings via Eq. \eqref{eq:gnn} and \eqref{eq:mlp};
	        \STATE Calculate the masked node and its contextual embeddings via Eq. \eqref{eq:gnn2}, \eqref{eq:readout}, and \eqref{eq:mlp2};
            \STATE Calculate the patch-level positive and negative scores for each node in $\mathbfcal{V}^U_B$ via Eq. \eqref{eq:patch-level anemone positive score} and \eqref{eq:patch-level anemone negative score};
            \STATE Calculate the patch-level extra negative scores for each node in $\mathbfcal{V}^L_B$ via Eq. \eqref{eq:patch-level anemone-fs negative score};
            \STATE Calculate the context-level positive and negative scores for each node in $\mathbfcal{V}^U_B$ via Eq. \eqref{eq:context-level anemone positive score} and \eqref{eq:context-level anemone negative score};
            \STATE Calculate the context-level extra negative scores for each node in $\mathbfcal{V}^L_B$ via Eq. \eqref{eq:context-level anemone-fs negative score};
            \STATE Calculate the loss $\mathcal{L}$ via Eq. \eqref{eq:patch-level loss}, \eqref{eq:context-level loss}, and \eqref{eq:loss};
			\STATE Back propagate to update trainable parameters $\mathbf{\Theta}$, $\mathbf{\Phi}$, $\mathbf{W}_p$, and $\mathbf{W}_c$;
			\ENDFOR
		\ENDFOR
		\STATE $/*$ {\it Model inference} $*/$
		\FOR{$v_i \in \mathbfcal{V}^U$}
		\FOR{evaluation round $r \in 1,2,\cdots,R$}
		\STATE Calculate $b^{(i)}_p$ and $b^{(i)}_c$ via Eq. \eqref{eq:base score};
		\ENDFOR
		\STATE Calculate the final patch-level and context-level anomaly scores $y^{(i)}_p$ and $y^{(i)}_c$ over $R$ evaluation rounds via Eq. \eqref{eq:final score v1};
		\STATE Calculate the final anomaly score $y^{(i)}$ of $v_i$ via Eq. \eqref{eq:final score v2};
		\ENDFOR
	\end{algorithmic}
\end{algorithm}

\subsection{Model Training and Algorithms} \label{subsec: optimization}

\noindent \textbf{Model training.} By combining the patch-level and context-level contrastive losses defined in Equation \ref{eq:patch-level loss} and \ref{eq:context-level loss}, we have the overall training objective by minimizing the following loss:

\begin{equation}
\mathcal{L}= \alpha \mathcal{L}_{c} + (1 - \alpha) \mathcal{L}_{p},
\label{eq:loss}
\end{equation}
where $\alpha$ is same as in Equation \ref{eq:final score v2} to balance the importance of two contrastive modules.

The overall procedures of \ourmethod and \ourmethodfs are in Algorithms \ref{algo: anemone} and \ref{algo: anemone-fs}. Specifically, in \ourmethod, we first sample a batch of nodes from the input attributed graph (line 5). Then, we calculate the positive and negative contrastive scores for each node (lines 6-10) to obtain the multi-scale contrastive losses, which is adopted to calculate the overall training loss (line 11) to update all trainable parameters (line 12). For \ourmethodfs, the differences are in two-folds. Firstly, it takes an attributed graph with a few available labeled anomalies as the input. Secondly, for labeled anomalies and unlabeled nodes in a batch, it has different contrastive routes in patch-level and context-level contrastive networks (lines 9-12). During the model inference, \ourmethod and \ourmethodfs shares the same anomaly scoring mechanism, where the statistical anomaly score for each node in $\mathbfcal{V}$ or$\mathbfcal{V}^U$ is calculated (lines 16-22 in Algorithm \ref{algo: anemone} and lines 18-24 in Algorithm \ref{algo: anemone-fs}
). 

\vspace{2mm}\noindent \textbf{Complexity analysis.} We analyse the time complexity of \ourmethod and \ourmethodfs algorithms in this subsection. For the shared anonymized subgraph sampling module, the time complexity of using RWR algorithm to sample a subgraph centred at $v_i$ is $\mathcal{O}(Kd)$, where $K$ and $d$ are the number of nodes in a subgraph and the average node degree in $\mathcal{G}$. Regarding the two proposed contrastive modules, their time complexities are mainly contributed by the underlying graph encoders, which are $\mathcal{O}(K^2)$. Thus, given $N$ nodes in $\mathbfcal{V}$, the time complexity of model training is $\mathcal{O}\big(NK(d+K)\big)$ in both \ourmethod and \ourmethodfs. During the model inference, the time complexity of \ourmethod is $\mathcal{O}\big(RNK(d+K)\big)$, where $R$ denotes the total evaluation rounds. For \ourmethodfs, its inference time complexity is $\mathcal{O}\big(RN^UK(d+K)\big)$, where $N^U = |\mathbfcal{V^U}|$ denotes the number of unlabeled nodes in $\mathcal{G}$.

\section{Experiments}
\label{sec:experiments}

In this section, we conduct a series of experiments to evaluate the anomaly detection performance of the proposed \ourmethod and \ourmethodfs on both unsupervised and few-shot learning scenarios. Specifically, we address the following research questions through experimental analysis:

\begin{table}[htbp]
	\centering
	\caption{The statistics of the datasets. The upper two datasets are social networks, and the remainders are citation networks.}
	\begin{tabular}{@{}c|c|c|c|c@{}}
		\toprule
		\textbf{Dataset}       &  \textbf{Nodes} & \textbf{Edges} & \textbf{Features} & \textbf{Anomalies} \\ 
		\midrule
		\textbf{Cora}  \cite{sen2008collective}        & 2,708           & 5,429          & 1,433             & 150                \\
		\textbf{CiteSeer}  \cite{sen2008collective}    & 3,327           & 4,732          & 3,703             & 150                \\
		\textbf{PubMed}   \cite{sen2008collective}     & 19,717          & 44,338         & 500               & 600                \\
		\textbf{ACM}  \cite{tang2008arnetminer}        & 16,484          & 71,980         & 8,337             & 600                \\ 
		\textbf{BlogCatalog} \cite{tang2009relational} & 5,196           & 171,743        & 8,189             & 300                \\
		\textbf{Flickr} \cite{tang2009relational}      & 7,575           & 239,738        & 12,407            & 450                \\
		\bottomrule
	\end{tabular}
	\label{table:dataset}
\end{table}

\begin{itemize}
    \item \textit{RQ1:} How do the proposed \ourmethod and \ourmethodfs perform in comparison to state-of-the-art graph anomaly detection methods? 
    \item \textit{RQ2:} How does the performance of \ourmethodfs change by providing different numbers of labeled anomalies?
    \item \textit{RQ3:} How do the contrastiveness in patch-level and context-level influence the performance of \ourmethod and \ourmethodfs?
    \item \textit{RQ4:} How do the key hyper-parameters impact the performance of \ourmethod?
\end{itemize}

\subsection{Datasets}

We conduct experiments on six commonly used datasets for graph anomaly detection, including four citation network datasets \cite{sen2008collective,tang2008arnetminer} (i.e., Cora, CiteSeer, PubMed, and ACM) and two social network datasets \cite{tang2009relational} (i.e., BlogCatalog and Flickr). Dataset statistics are summarized in Table \ref{table:dataset}.

Since ground-truth anomalies are inaccessible for these datasets, we follow previous works \cite{dominant_ding2019deep,cola_liu2021anomaly} to inject two types of synthetic anomalies (i.e., structural anomalies and contextual anomalies) into the original graphs. For structural anomaly injection, we use the injection strategy proposed by \cite{anoinj_s_ding2019interactive}: several groups of nodes are randomly selected from the graph, and then we make the nodes within one group fully linked to each other. In this way, such nodes can be regarded as structural anomalies. To generate contextual anomalies, following \cite{anoinj_c_song2007conditional}, a target node along with $50$ auxiliary nodes are randomly sampled from the graph. Then, we replace the features of the target node with the features of the farthest auxiliary node (i.e., the auxiliary node with the largest features' Euclidean distance to the target node). By this, we denote the target node as a contextual anomaly. We inject two types of anomalies with the same quantity and the total number is provided in the last column of Table \ref{table:dataset}.

\subsection{Baselines}

We compare our proposed \ourmethod and \ourmethodfs with three types of baseline methods, including (1) shallow learning-based unsupervised methods (i.e., AMEN \cite{amen_perozzi2016scalable}, Radar \cite{radar_li2017radar}), (2) deep learning-based unsupervised methods (i.e., ANOMALOUS \cite{anomalous_peng2018anomalous}, DOMINANT \cite{dominant_ding2019deep}, and CoLA \cite{cola_liu2021anomaly}), and (3) semi-supervised methods (i.e., DeepSAD \cite{deepsad_ruff2019deep}, SemiGNN \cite{semignn_wang2019semi}, and GDN \cite{gdn_ding2021few}). Details of these methods are introduced as following: 
\begin{itemize}
    \item \textbf{AMEN} \cite{amen_perozzi2016scalable} is an unsupervised graph anomaly detection method which detects anomalies by analyzing the attribute correlation of ego-network of nodes.
    
    \item \textbf{Radar} \cite{radar_li2017radar} identifies anomalies in graphs by residual and attribute-structure coherence analysis.
    
    \item \textbf{ANOMALOUS} \cite{anomalous_peng2018anomalous} is an unsupervised method for attributed graphs, which performs anomaly detection via CUR decomposition and residual analysis.
    
    \item \textbf{DGI} \cite{dgi_velickovic2019deep} is an unsupervised contrastive learning method for representation learning. In DGI, we use the score computation module in \cite{cola_liu2021anomaly} to estimate nodes' abnormality.
    
    \item \textbf{DOMINANT} \cite{dominant_ding2019deep} is a deep graph autoencoder-based unsupervised method that detects anomalies by evaluating the reconstruction errors of each node.
    
    \item \textbf{CoLA} \cite{cola_liu2021anomaly} is a contrastive learning-based anomaly detection method which captures anomalies with a GNN-based contrastive framework.
    
    \item \textbf{DeepSAD} \cite{deepsad_ruff2019deep} is a deep learning-based anomaly detection method for non-structured data. We take node attributes as the input of DeepSAD.
    
    \item \textbf{SemiGNN} \cite{semignn_wang2019semi} is a semi-supervised fraud detection method that use attention mechanism to model the correlation between different neighbors/views.
    
    \item \textbf{GDN} \cite{gdn_ding2021few} is a GNN-based model that detects anomalies in few-shot learning scenarios. It leverages a deviation loss to train the detection model in an end-to-end manner.
\end{itemize}

\subsection{Experimental Setting}

\mysubsubtitle{Evaluation Metric}
We employ a widely used metric, AUC-ROC \cite{dominant_ding2019deep,cola_liu2021anomaly}, to evaluate the performance of different anomaly detection methods. The ROC curve indicates the plot of true positive rate against false positive rate, and the AUC value is the area under the ROC curve. The value of AUC is within the range $[0,1]$ and a larger value represents a stronger detection performance. To reduce the bias caused by randomness and compare fairly \cite{cola_liu2021anomaly}, for all datasets, we conduct a $5$-run experiment and report the average performance.\\ 

\begin{figure*}[ht]
	\centering
		\subfigure[ROC curve of Cora.]{
		\includegraphics[width=.45\linewidth]{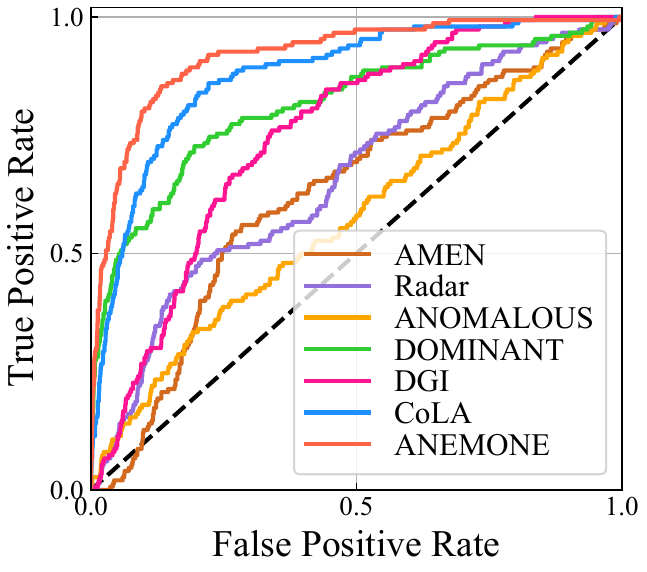}
	}\hspace{-2mm}
			\subfigure[ROC curve of CiteSeer.]{
		\includegraphics[width=.45\linewidth]{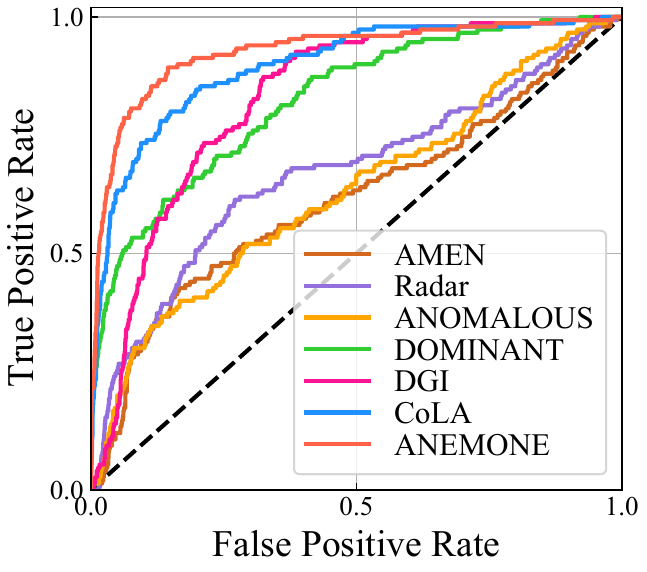}
	}\hspace{-2mm}
		\subfigure[ROC curve of ACM.]{
		\includegraphics[width=.45\linewidth]{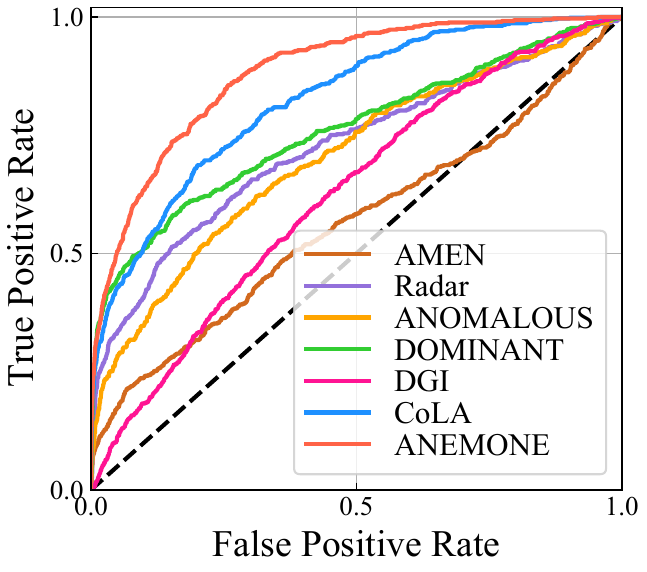}
	}\hspace{-2mm}
		\subfigure[ROC curve of BlogCatalog.]{
		\includegraphics[width=.45\linewidth]{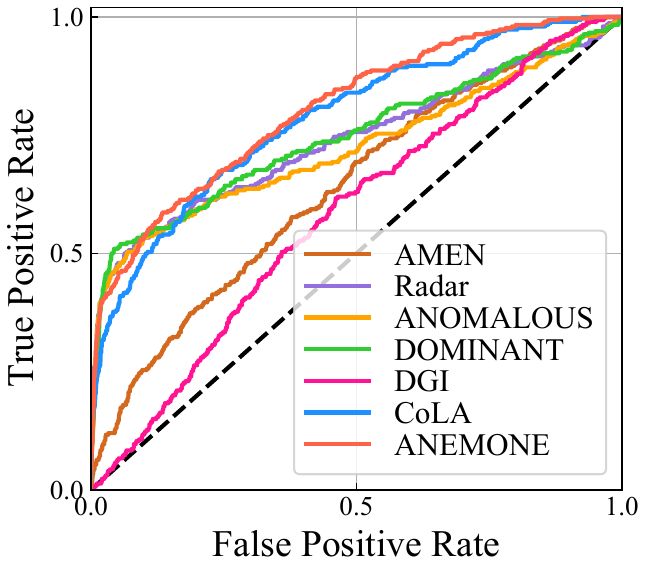}
		\label{subfig:parameter}
	}
	\caption{The comparison of ROC curves on four datasets in unsupervised learning scenario.}
	\label{fig:roc}
\end{figure*}

\mysubsubtitle{Dataset Partition}
In an unsupervised learning scenario, we use graph data $\mathcal{G}=(\mathbf{X},\mathbf{A})$ to train the models, and evaluate the anomaly detection performance on the full node set $\mathbfcal{V}$ (including all normal and abnormal nodes). 
In few-shot learning scenario, we train the models with $\mathcal{G}$ and $k$ anomalous labels (where $k$ is the size of labeled node set $|\mathbfcal{V}^L|$), and use the rest set of nodes $\mathbfcal{V}^U$ to measure the models' performance. \\

\mysubsubtitle{Parameter Settings}
In our implementation, the size $K$ of subgraph and the dimension of embeddings are fixed to $4$ and $64$, respectively. The trade-off parameter $\alpha$ is searched in $\{0.2, 0.4, 0.6, 0.8, 1\}$. The number of testing rounds of anomaly estimator is set to $256$. We train the model with Adam optimizer with a learning rate $0.001$. For Cora, Citeseer, and Pubmed datasets, we train the model for $100$ epochs; for ACM, BlogCatalog, and Flickr datasets, the numbers of epochs are $2000$, $1000$, and $500$, respectively. 

\begin{table*}[htbp]
	\small
	\centering
	\caption{The comparison of anomaly detection performance (i.e., AUC) in unsupervised learning scenario. The best performance is highlighted in \textbf{bold}.}
	{
    	\begin{tabular}{p{85 pt}<{}|p{45 pt}<{\centering}p{45 pt}<{\centering}p{45 pt}<{\centering}p{45 pt}<{\centering}p{45 pt}<{\centering}p{45 pt}<{\centering}}
    		\toprule
    		Method    & Cora            & CiteSeer        & PubMed     & ACM                & BlogCatalog     & Flickr              \\
    		\midrule
			AMEN \cite{amen_perozzi2016scalable}  & 0.6266          & 0.6154          & 0.7713             & 0.5626          & 0.6392          & 0.6573               \\
			Radar \cite{radar_li2017radar}    & 0.6587          & 0.6709          & 0.6233       & 0.7247        & 0.7401          & 0.7399                 \\
			ANOMALOUS \cite{anomalous_peng2018anomalous}  & 0.5770          & 0.6307          & 0.7316        & 0.7038      & 0.7237          & 0.7434                \\
			\midrule
			DGI  \cite{dgi_velickovic2019deep}   & 0.7511          & 0.8293          & 0.6962     & 0.6240         & 0.5827          & 0.6237              \\
			DOMINANT \cite{dominant_ding2019deep}    & 0.8155          & 0.8251          & 0.8081    & 0.7601      & 0.7468          & 0.7442                   \\
			CoLA  \cite{cola_liu2021anomaly}   & 0.8779          & 0.8968          & 0.9512       & 0.8237      & 0.7854          & 0.7513  \\
    		\midrule
    		\ourmethod       & \textbf{0.9057} & \textbf{0.9189} & \textbf{0.9548} & \textbf{0.8709} & \textbf{0.8067} & \textbf{0.7637} \\
    		\bottomrule
    	\end{tabular}
    }
\label{table:overall_unsup}
\end{table*}

\subsection{Performance Comparison (RQ1)}

We evaluate \ourmethod in unsupervised learning scenario where labeled anomaly is unavailable, and evaluate \ourmethodfs in few-shot learning scenario ($k=10$) where $10$ annotated anomalies are known during model training. \\

\begin{table*}[htbp]
	\small
	\centering
	\caption{The comparison of anomaly detection performance (i.e., AUC) in few-shot learning scenario. The best performance is highlighted in \textbf{bold}.}
	{
    	\begin{tabular}{p{85 pt}<{}|p{45 pt}<{\centering}p{45 pt}<{\centering}p{45 pt}<{\centering}p{45 pt}<{\centering}p{45 pt}<{\centering}p{45 pt}<{\centering}}
    		\toprule
    		Method    & Cora            & CiteSeer        & PubMed     & ACM                & BlogCatalog     & Flickr              \\
    		\midrule
			AMEN \cite{amen_perozzi2016scalable}  & 0.6257&0.6103&0.7725&0.5632&0.6358&0.6615\\
			Radar \cite{radar_li2017radar}    &0.6589&0.6634&0.6226&0.7253&0.7461&0.7357\\
			ANOMALOUS \cite{anomalous_peng2018anomalous}  &0.5698&0.6323&0.7283&0.6923&0.7293&0.7504\\
			\midrule
			DGI  \cite{dgi_velickovic2019deep}   &0.7398&0.8347&0.7041&0.6389&0.5936&0.6295\\
			DOMINANT \cite{dominant_ding2019deep}    &0.8202&0.8213&0.8126&0.7558&0.7391&0.7526\\
			CoLA  \cite{cola_liu2021anomaly}   &0.8810&0.8878&0.9517&0.8272&0.7816&0.7581\\
    		\midrule
    		DeepSAD  \cite{deepsad_ruff2019deep} & 0.4909&	0.5269&	0.5606&	0.4545&	0.6277&	0.5799\\
			SemiGNN \cite{semignn_wang2019semi}    & 0.6657&	0.7297&	OOM&	OOM&	0.5289&	0.5426\\
			GDN  \cite{gdn_ding2021few}   &0.7577&0.7889&0.7166&0.6915&0.5424&0.5240\\
    		\midrule
    		\ourmethod       & {0.8997} & {0.9191} & {0.9536} & {0.8742} & {0.8025} & {0.7671} \\
    		\ourmethodfs       & \textbf{0.9155} & \textbf{0.9318} & \textbf{0.9561} & \textbf{0.8955} & \textbf{0.8124} & \textbf{0.7781} \\
    		\bottomrule
    	\end{tabular}
    }
\label{table:overall_fs}
\end{table*}

\mysubsubtitle{Performance in Unsupervised Learning Scenario}
In unsupervised scenario, we compared \ourmethod with $6$ unsupervised baselines. The ROC curves on $4$ representative datasets are illustrated in Fig. \ref{fig:roc}, and the comparison of AUC value on all $6$ datasets is provided in Table \ref{table:overall_unsup}. From these results, we have the following observations.
\begin{itemize}
    \item \ourmethod consistently outperforms all baselines on six benchmark datasets. The performance gain is due to (1) the two-level contrastiveness successfully capturing anomalous patterns in different scales and (2) the well-designed anomaly estimator effectively measuring the abnormality of each node.
    \item The deep learning-based methods significantly outperform the shallow learning-based methods, which illustrates the capability of GNNs in modeling data with complex network structures and high-dimensional features.
    \item The ROC curves by \ourmethod are very close to the points in the upper left corner, indicating our method can precisely discriminate abnormal samples from a large number of normal samples.
\end{itemize}


\mysubsubtitle{Performance in Few-shot Learning Scenario}
In few-shot learning scenario, we consider both unsupervised and semi-supervised baseline methods for the comparison with our methods. The results are demonstrated in Table \ref{table:overall_fs}. As we can observe, \ourmethod and \ourmethodfs achieve consistently better performance than all baselines, which validates that our methods can handle few-shot learning setting as well. Also, we find that \ourmethodfs has better performance than \ourmethod, meaning that our proposed solution for few-shot learning can further leverage the knowledge from a few numbers of labeled anomalies. In comparison, the semi-supervised learning methods (i.e., DeepSAD, SemiGNN, and GDN) do not show a competitive performance, indicating their limited capability in exploiting the label information.

\subsection{Few-shot Performance Analysis (RQ2)}


In order to verify the effectiveness of \ourmethodfs in different few-shot anomaly detection settings, we change the number $k$ of anomalous samples for model training to form $k$-shot learning settings for evaluation. We perform experiments on four datasets (i.e., Cora, CiteSeer, ACM, and BlogCatalog) and select $k$ from $\{1,3,5,10,15,20\}$. The experimental results are demonstrated in Table \ref{table:fewshot} where the performance in unsupervised setting (denoted as ``Unsup.'') is also reported as a baseline.

\begin{table}[htbp]
	\small
	\centering
	\caption{Few-shot performance analysis of \ourmethodfs.}
	{
    	\begin{tabular}{l|cccc}
    		\toprule
    		Setting    & Cora            & CiteSeer       & ACM                & BlogCatalog   \\
    		\midrule
    		Unsup. & 0.8997&0.9191&0.8742&0.8025\\
    		\midrule
			1-shot  &0.9058&0.9184&0.8858&0.8076\\
			3-shot  &0.9070&0.9199&0.8867&0.8125\\
			5-shot  &0.9096&0.9252&0.8906&0.8123\\
			10-shot &0.9155&0.9318&0.8955&0.8124\\
			15-shot &0.9226&0.9363&0.8953&0.8214\\
			20-shot &0.9214&0.9256&0.8965&0.8228\\
    		\bottomrule
    	\end{tabular}
    }
\label{table:fewshot}
\end{table}

\begin{figure}[t]
	\centering
		\subfigure[AUC w.r.t. $\alpha$ in \ourmethod.]{
		\includegraphics[width=.45\linewidth]{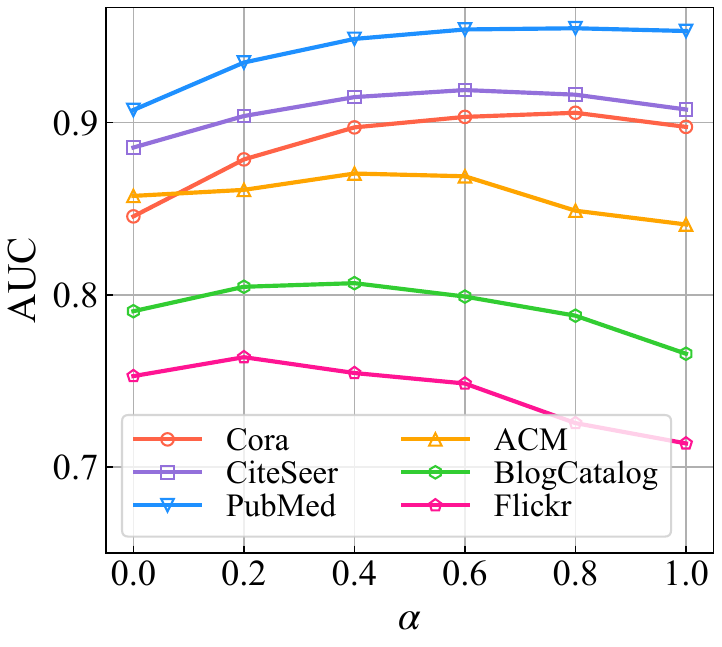}
	}\hspace{-2mm}
			\subfigure[AUC w.r.t. $\alpha$ in \ourmethodfs.]{
		\includegraphics[width=.45\linewidth]{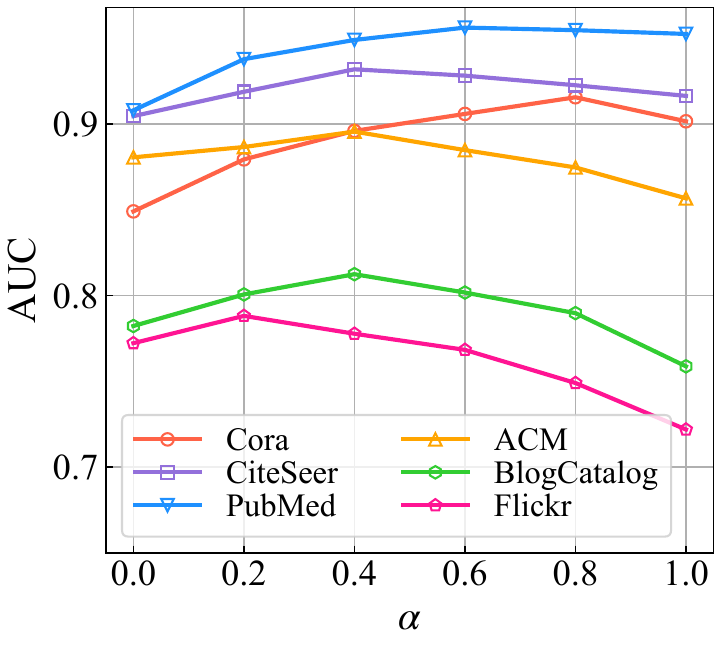}
	}
	\caption{Anomaly detection performance with different selection of trade-off parameter $\alpha$ in unsupervised and few-shot learning scenarios.}
	\label{fig:ablation}
\end{figure}

\begin{figure*}[ht]
	\centering
		\subfigure[AUC w.r.t. number of evaluation rounds.]{
		\includegraphics[width=.45\linewidth]{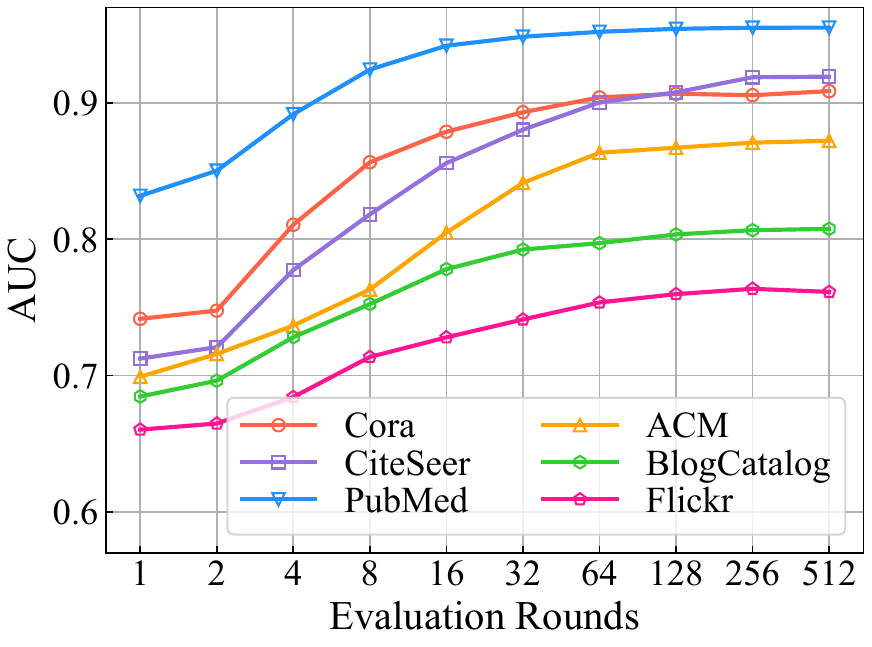}\label{subfig:round}
	}\hfill
			\subfigure[AUC w.r.t. size of subgraph.]{
		\includegraphics[width=.45\linewidth]{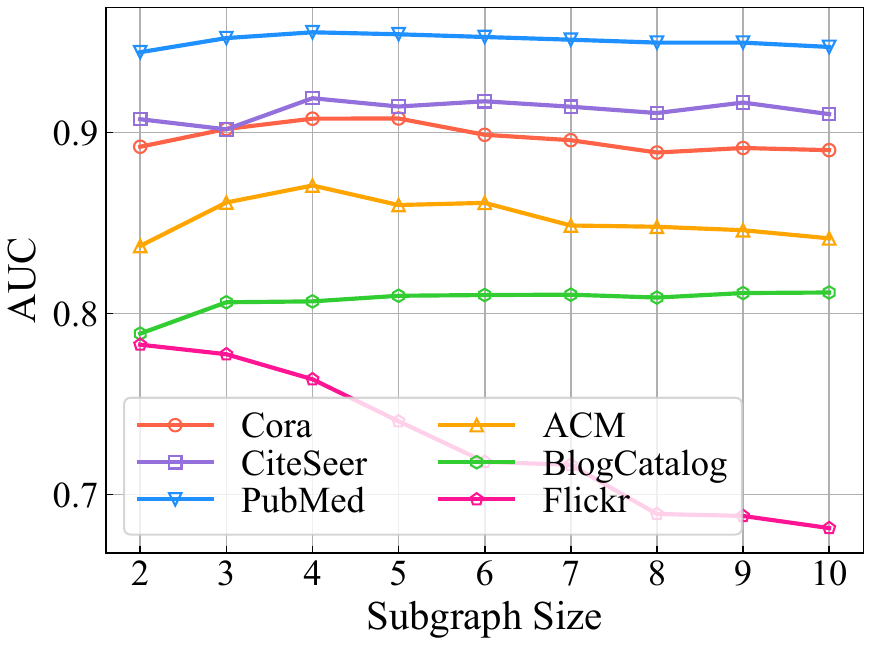}\label{subfig:subg}
	}\hfill
		\subfigure[AUC w.r.t. hidden dimension.]{
		\includegraphics[width=.45\linewidth]{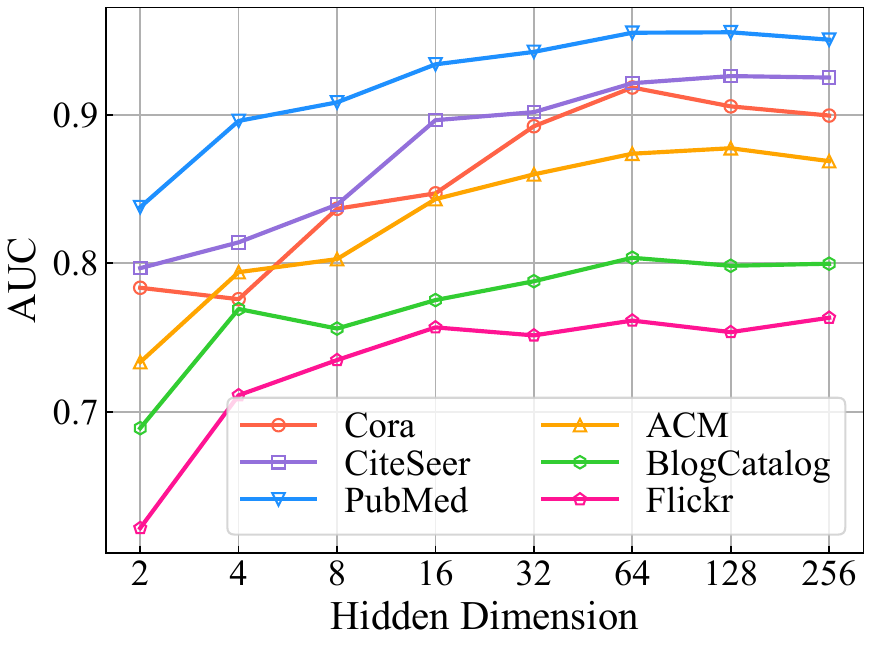}\label{subfig:dim}
	}
	\caption{Parameter sensitivities of \ourmethod w.r.t. three hyper-parameters on six benchmark datasets.}
	\label{fig:param}
\end{figure*}

The results show that the \ourmethodfs can achieve good performance even when only one anomaly node is provided (i.e., 1-shot setting). A representative example is the results of ACM dataset where a $1.16\%$ performance gain is brought by one labeled anomaly. Such an observation indicates that \ourmethodfs can effectively leverage the knowledge from scarce labeled samples to better model the anomalous patterns. Another finding is that the anomaly detection performance generally increases following the growth of $k$ especially when $k\leq15$. This finding demonstrates that \ourmethodfs can further optimize the anomaly detection model when more labeled anomalies are given.

\subsection{Ablation Study (RQ3)}

In this experiment, we investigate the contribution of patch- and context- level contrastiveness to the anomaly detection performance of \ourmethod and \ourmethodfs. In concrete, we adjust the value of trade-off parameter $\alpha$ and the results are illustrated in Fig. \ref{fig:ablation}. Note that $\alpha = 0$ and $\alpha = 1$ mean that the model only considers patch- and context level contrastive learning, respectively.

As we can observe in Fig. \ref{fig:ablation}, \ourmethod and \ourmethodfs can achieve the highest AUC values when $\alpha$ is between $0.2$ and $0.8$, and the best selections of $\alpha$ for each dataset are quite different. Accordingly, we summarize that jointly considering the contrastiveness in both levels always brings the best detection performance. We also notice that in some datasets (i.e., Cora, CiteSeer, and PubMed) context-level contrastive network performs better than the patch-level one, while in the rest datasets the patch-level contrastiveness brings better results. It suggests that two types of contrastiveness have unique contributions in identifying anomalies from network data with diverse properties.

\subsection{Parameter Sensitivity (RQ4)}

To study how our method is impacted by the key hyper-parameters, we conduct experiments for \ourmethod with different selections of evaluation rounds $R$, subgraph size $K$, and hidden dimension $D'$. 

\vspace{2mm}\mysubsubtitle{Evaluation Rounds}
To explore the sensitivity of \ourmethod to evaluation rounds $R$, we tune the $R$ from $1$ to $512$ on six datasets, and the results are demonstrated in Fig. \ref{subfig:round}. As we can find in the figure, the detection performance is relatively poor when $R<4$, indicating that too few evaluation rounds are insufficient to represent the abnormality of each node. When $R$ is between $4$ and $256$, we can witness a significant growing trade of AUC following the increase of $R$, which demonstrates that adding evaluation rounds within certain ranges can significantly enhance the performance of \ourmethod. An over-large $R$ ($R=512$) does not boost the performance but brings heavier computational cost. Hence, we fix $R=256$ in our experiments to balance performance and efficiency. 

\vspace{2mm}\mysubsubtitle{Subgraph Size}
In order to investigate the impact of subgraph size, we search the node number of contextual subgraph $K$ in the range of $\{2,3,\cdots,10\}$. We plot the results in Fig. \ref{subfig:subg}. We find that \ourmethod is not sensitive to the choice of $K$ on datasets except Flickr, which verifies the robustness of our method. For citation networks (i.e., Cora, CiteSeer, PubMed, and ACM), a suitable subgraph size between $3$ and $5$ results in the best performance. Differently, BlogCatalog requires a larger subgraph to consider more contextual information, while Flickr needs a smaller augmented subgraph for contrastive learning. 

\vspace{2mm}\mysubsubtitle{Hidden Dimension}
In this experiment, we study the selection of hidden dimension $D'$ in \ourmethod. We alter the value of $D'$ from $2$ to $256$ and the effect of $D'$ on AUC is illustrated in Fig. \ref{subfig:dim}. As shown in the figure, when $D'$ is within $[2,64]$, there is a significant boost in anomaly detection performance with the growth of $D'$. This observation indicates that node embeddings with a larger length can help \ourmethod capture more complex information. We also find that the performance gain becomes light when $D'$ is further enlarged. Consequently, we finally set $D'=64$ in our main experiments.

\section{Conclusion}
\label{sec:conclusion}
In this paper, we investigate the problem of graph anomaly detection. By jointly capturing anomalous patterns from multiple scales with both patch level and context level contrastive learning, we propose a novel algorithm, \ourmethod, to learn the representation of nodes in a graph. With a statistical anomaly estimator to capture the agreement from multiple perspectives, we predict an anomaly score for each node so that anomaly detection can be conducted subsequently. As a handful of ground-truth anomalies may be available in real applications, we further extend our method as \ourmethodfs, a powerful method to utilize labeled anomalies to handle the settings of few-shot graph anomaly detection. Experiments on six benchmark datasets validate the performance of the proposed \ourmethod and \ourmethodfs.  In the future, leverage the in-context learning capacity which only requires a few samples, we will look into foundation models or large language models \cite{pan2024integrating,liu2024arc}, for effective graph anomaly detection,

\bibliography{reference}

\end{document}